\pdfoutput=1
\documentclass[11pt]{article}

\usepackage[]{acl}

\usepackage{times}
\usepackage{latexsym}
\usepackage{amsmath}
\usepackage{booktabs}
\usepackage{multirow}
\usepackage{pifont}
\usepackage{graphicx}  % per \rotatebox
\usepackage{arydshln}  % per \cdashline
\usepackage[T1]{fontenc}
\usepackage[utf8]{inputenc}

\usepackage{microtype}

\usepackage{inconsolata}
\usepackage{xcolor}%
\usepackage{textcomp}%
\usepackage{manyfoot}%
\usepackage{booktabs}%
\usepackage{algorithm}%
\usepackage{algorithmicx}%
\usepackage{algpseudocode}%
\usepackage{listings}%
\usepackage{colortbl}
\usepackage{booktabs}
\usepackage{float}
\usepackage{xcolor}
\usepackage{lineno}
\usepackage{booktabs}
\usepackage{array}
\usepackage{longtable}
\usepackage{pdflscape}
\usepackage{graphicx}
\definecolor{mygray}{gray}{0.8}
\definecolor{mygray2}{gray}{.95}
\definecolor{verdescuro}{rgb}{0.3,.7,0.3}
\definecolor{verdechiaro}{rgb}{0.6,.9,0.6}
\definecolor{lightblue}{rgb}{0.68, 0.85, 0.9}
\definecolor{lightindianred}{rgb}{0.87, 0.58, 0.58}
\DeclareSymbolFont{extraup}{U}{zavm}{m}{n}
\DeclareMathSymbol{\newcheckmark}{\mathalpha}{extraup}{128}%uni2713
\DeclareMathSymbol{\newcrossmark}{\mathalpha}{extraup}{129}%uni2717

\title{
Are LLMs effective psychological assessors? Leveraging adaptive RAG for interpretable mental health screening 
through psychometric practice
}
\author{Federico Ravenda$^{1,2}$, Seyed Ali Bahrainian$^{3,4}$, \\  \textbf{Andrea Raballo}$^{1}$\textbf{,} \textbf{Antonietta Mira}$^{1,5}$\textbf{,} \textbf{Noriko Kando}$^{2}$\\
 \texttt{federico.ravenda@usi.ch},
  \texttt{bahrainian@brown.edu} \\ \texttt{andrea.raballo@usi.ch}, \texttt{antonietta.mira@usi.ch},
  \texttt{kando@nii.ac.jp}\\
  $^1$Euler Institute, Università della Svizzera italiana,
  $^2$National Institute of Informatics\\
  $^3$Brown University,
  $^4$University of Tübingen,
  $^5$Insubria University\\
}

\begin{document}
\maketitle
\begin{abstract}
In psychological practices, standardized questionnaires serve as essential tools for assessing mental health through structured, clinically-validated questions (i.e., items). While social media platforms offer rich data for mental health screening, computational approaches often bypass these established clinical assessment tools in favor of black-box classification.
We propose a novel questionnaire-guided screening framework that bridges psychological practice and computational methods through adaptive Retrieval-Augmented Generation (\textit{aRAG}). Our approach links unstructured social media content and standardized clinical assessments by retrieving relevant posts for each questionnaire item and using Large Language Models (LLMs) to complete validated psychological instruments. 
Our findings demonstrate two key advantages of questionnaire-guided screening: First, when completing the Beck Depression Inventory-II (BDI-II), our approach matches or outperforms state-of-the-art performance on Reddit-based benchmarks without requiring training data. Second, we show that guiding LLMs through standardized questionnaires can yield superior results compared to directly prompting them for depression screening, while also providing a more interpretable assessment by linking model outputs to clinically validated diagnostic criteria. Additionally, we show, as a proof-of-concept, how our questionnaire-based methodology can be extended to other mental conditions' screening, 
highlighting the promising role of LLMs as psychological assessors.
\footnote{ \small Code available: \href{https://github.com/Fede-stack/Adaptive-RAG-for-Psychological-Assessment}{https://github.com/Fede-stack/Adaptive-RAG-for-Psychological-Assessment}\normalsize}

\end{abstract}

\section{Introduction}

According to the World Health Organization (WHO), one in seven adolescents experiences a mental health disorder~\cite{wiederhold2022escalating}, with depression, anxiety, and behavioral disorders leading among young people. Following COVID-19, mental health conditions (MHCs) surged, with depressive disorders increasing by 28\% in 2020~\cite{kieling2011child,winkler2020increase}.
Given the extent of this need, the WHO Special Initiative for Mental Health prioritizes improving and expanding access to quality mental health interventions and services as a key strategic goal \cite{WHO2022}.

Psychological questionnaires play a crucial role in describing mental states by measuring various psychological constructs, 
as outlined in the Diagnostic and Statistical Manual of Mental Disorders (DSM)~\cite{hopwood2012dsm}. 
The conventional interpretation of data derived from psychometric scales assumes that obtained scores reflect the intensity of a respondent attitudes. Psychological questionnaires can be used to assess various constructs related to different mental disorders as screening methods to develop an initial clinical profile. 

In this work, we focus on different widely used standardized psychological questionnaires, specifically the Beck Depression Inventory-II (BDI-II)~\cite{beck1996beck} for depression screening, the Self-Harm Inventory (SHI)~\cite{sansone2010measuring} for self-harm behaviour detection, the SCOFF questionnaire~\cite{morgan1999scoff} for eating disorders screening, and the Pathological Gambling Diagnostic Form from DSM-V~\cite{hopwood2012dsm}. All of them are self-reported surveys where overall scores correspond to specific severity levels of the respective conditions.

Despite their clinical validity, traditional psychological questionnaires present several practical limitations. They typically require in-person administration by trained professionals, making large-scale screening logistically challenging and resource-intensive. Access barriers disproportionately affect underserved populations, creating equity concerns in mental health assessment. 

Increasingly, people are turning to social networks as a space to express their feelings and experiences and find support~\cite{bucci2019digital,naslund2016future}. Numerous initiatives have emerged to analyze social media content for health monitoring using NLP techniques, including CLPsych~\cite{tsakalidis2022overview} and eRisk~\cite{losada2017erisk}, through organized completion tasks. 

Recent research has revealed significant limitations in using closed-source LLMs for mental disorder classification. Studies by \cite{amin2023will} and \cite{xu2023leveraging} demonstrate that both zero-shot and few-shot approaches struggle to match state-of-the-art (SOTA) supervised methods. These limitations stem from several challenges: (1) LLMs' difficulty in directly mapping unstructured text to diagnostic categories, (2) the complex, multi-dimensional nature of mental health assessment requiring domain expertise, and (3) the semantic gap between social media language and clinical criteria. We propose mitigating these challenges through an intermediate step: rather than attempting direct diagnosis, we instruct LLMs to complete standardized psychological questionnaires, effectively decomposing the complex diagnostic task into structured, clinically-validated assessment items.

To achieve that, we propose an adaptive RAG approach (aRAG), combining retrieval and classification, to accurately predict users' responses to psychological questionnaire items by analyzing their Reddit post history. Unlike existing methods that use fixed retrieval parameters or direct classification, our approach automatically determines the optimal number of relevant posts needed for each questionnaire item, adapting to the semantic density and relevance of available content. To the best of our knowledge, we demonstrate, for the first time, how  LLMs (both open- and closed-source) can serve as effective annotators of standardized psychological questionnaires by analyzing social media posts through aRAG. 

The main contributions of this paper are:
\textbf{(1)} We explore various combinations of open- and closed-source LLMs together with dense retrievers for predicting psychological questionnaires, evaluating how performance varies with different combinations of LLMs, prompt strategies, and retrieval models.
\textbf{(2)} We compare the results of our approach with the best results obtained for the considered eRisk collections using primarily supervised models, showing how our unsupervised approach often outperforms the benchmarks.
\textbf{(3)} We extend this paradigm to other mental disorders, introducing an interpretable and unsupervised method for predicting new MHCs.

These results confirm that LLMs can serve as effective and promising psychological assessors when their predictions are guided by standardized clinical instruments, bridging the gap between language models and psychometric practice.

\section{Related Works} \label{rel}

Recent advancements in NLP have enabled the development of new and complex models across various areas, particularly in digital and mental health~\cite{rissola2019anticipating}. 
Transformer-based models \cite{vaswani2017attention} have significantly advanced mental health analysis on social media platforms. While initial work used BERT variants for depression detection \cite{raj2024depression,rissola2020dataset}, specialized models like MentalBERT \cite{ji2022mentalbert} emerged, pre-trained specifically on mental health-related content. A novel direction explored how emotion manifests in depressed individuals' \cite{bucur2022life} and a broader spectrum of mental disorders \cite{de2024emotional} from social media posts, using emotional and psychological markers to provide interpretable assessment. 

Recently, Large Language Models have shown increasing promise, with approaches like MentaLLaMA \cite{yang2024mentallama} offering interpretable analysis, and \cite{varadarajan2024archetypes} combining theoretical frameworks with computational techniques for suicide risk assessment. These advances suggest potential for real-time intervention and support \cite{yang2024behavioral}.

Concerning the use of NLP methods to predict psychological questionnaires' responses, early approaches  used neural models to predict personality traits \cite{elourajini2022aws} and Myers-Briggs indicators \cite{yang2021learning} from user-generated social media content. BERT embeddings have been leveraged to link social media expressions with psychological assessments \cite{vu2020predicting,atari2023contextualized}. More recently, \cite{rosenman2024llm} used an LLM to impersonate interviewees and complete questionnaires, using these responses as features in a Random Forest to predict new questionnaire scores.
\begin{figure*}[ht]
    \centering
    \includegraphics[ width=.9\linewidth]{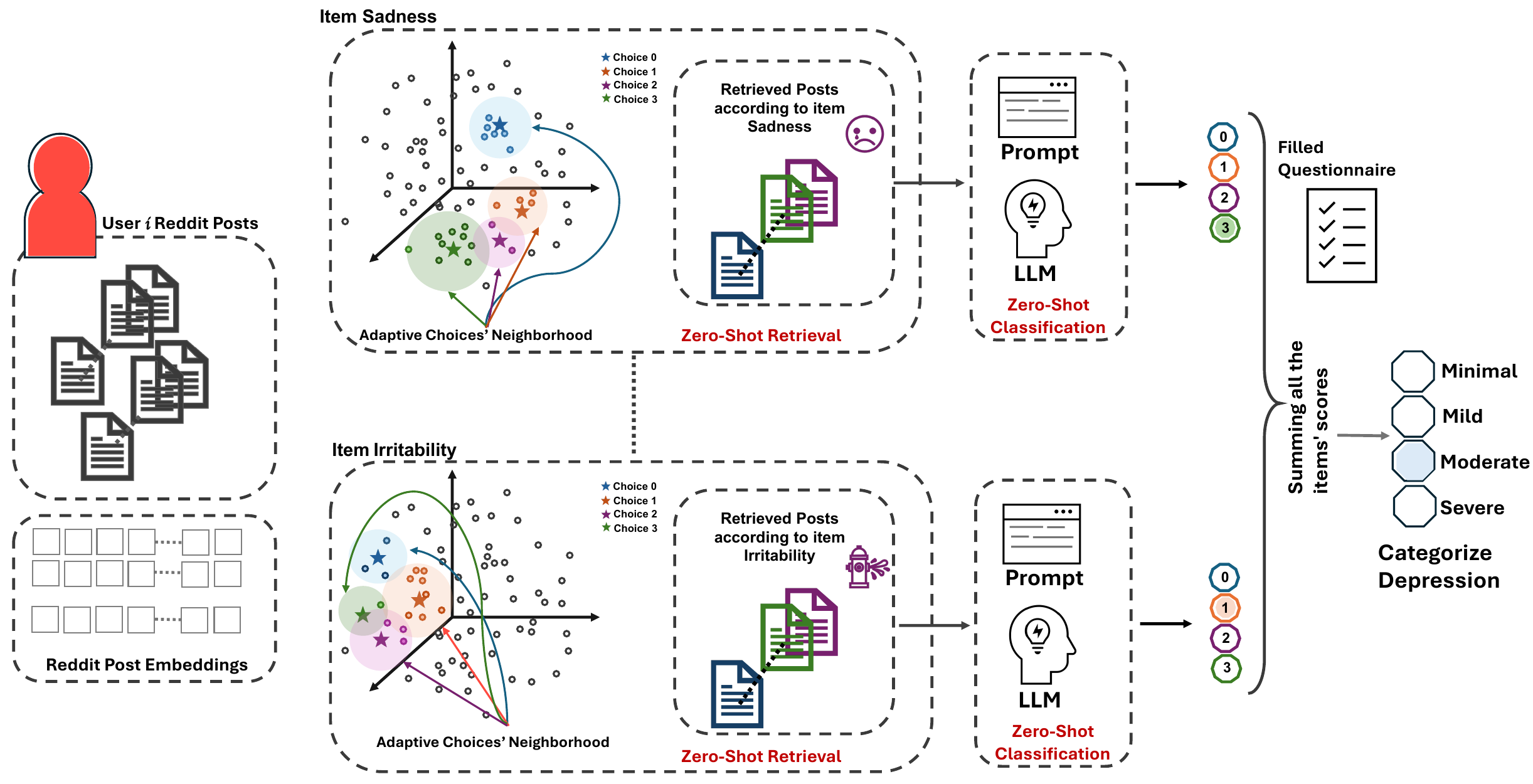}
    \caption{Pipeline of the main steps of our architecture. For each user, embeddings are created for each post and for each of the 4 choices of each item. The most relevant posts for each choice are retrieved and used as input for the LLM to generate the item score.}
    \label{steps}
\end{figure*}

With regard to eRisk data, recent approaches were used to predict BDI-II responses using advanced computational methods. \cite{perez2023bdi} introduced a retrieval-based framework with item-based classifiers for depression screening, while \cite{ravenda2024} proposed a probabilistic approach combining retrieval mechanisms to handle ordinal Likert scales. Both approaches innovate via retrieval-based selection of relevant social media content, departing from previous methods that used fixed post selection.

Our work differs from previous approaches by focusing on a completely unsupervised scenario, leveraging LLMs through an aRAG approach, filtering the most relevant posts and use LLMs to predict scores, establishing a semantic mappings between social media content and standardized questionnaire items. To evaluate the effectiveness of our approach in such tasks, we use two eRisk collections from the 2019 and 2020 editions~\cite{losada2019overview,losada2020erisk}, which contain the post-history of Reddit users alongside their completed BDI-II questionnaires. After demonstrating the effectiveness of this unsupervised approach, we extend it to different MHCs, highlighting the benefits of constraining LLM predictions to individual psychological questionnaire responses.

\section{Research Questions}

\noindent The main \textbf{Research Questions} we address are:

\noindent\textbf{(RQ1.)} Is it possible to fill out a psychological questionnaire automatically based on a user's Reddit post history in an unsupervised context? How does this compare in terms of performance to SOTA models for the considered datasets?\\
\noindent\textbf{(RQ2.)} How does the model's effectiveness vary with changes in:\\
\textbf{(RQ2a.)} The LLM being used. To answer this question, we consider six LLMs: two large-scale open-source (\texttt{Qwen 2.5 70B}, \texttt{DeepSeek V3}), two lightweight open-source (\texttt{Phi-3-mini}, \texttt{Phi-3.5-mini}), and 2 closed-source (\texttt{Claude-3.5-Sonnet}, \texttt{gpt-4o-mini}).\\
\textbf{(RQ2b.)} The prompting strategy we employ. Specifically, we use both a direct prompting approach and Chain-of-Thought (CoT) \cite{wei2022chain} prompting.\\
\textbf{(RQ2c.)} The dense retrieval models. These are used to retrieve the most relevant posts for each user in response to each survey item.\\
\noindent\textbf{(RQ3.)} Can our approach -completing a standardized psychological questionnaire
to obtain a ``psychological explanation'' of why a user is associated with certain risk levels - be extended to
other MHCs without training data? 
\\
\noindent\textbf{(RQ4.)} What are the benefits of our \textit{psychological-guided} approach compared to an approach that relies exclusively on prompting?\\

\normalsize

\noindent \textbf{Problem Definition.} 
While existing computational approaches typically frame mental health prediction as a direct mapping from text to diagnosis \( f(\text{Text}) \to Y \) \cite{kim2020deep,sekulic2019adapting}, where Text represents textual information like social media posts or interview transcriptions, this black-box formulation faces significant challenges in clinical applicability, interpretability, and generalizability across contexts \cite{paris2012differences,friginal2017linguistic}. These models, often neural, require large datasets to perform well.
Our work introduces a novel paradigm that bridges computational and clinical practice by leveraging standardized psychological questionnaires as an intermediate structured representation. By reformulating the prediction task through questionnaire items, our method offers several key advantages: (1) clinical interpretability through standardized assessment criteria, (2) transparent reasoning through item-level predictions, and (3) alignment with established psychological practice. The prediction is framed as a function linking text and questionnaire items, \( f(\text{Text}, \text{Item}_i) \to S_i \), where \( S_i \) represents the user's score for item \( i \). The combination of these individual scores defines a final score used to diagnose symptoms as \( \sum_i f(\text{Text}, \text{Item}_i) \to Y \). 
The proposed method comprises two main steps, illustrated in Figure \ref{steps}: \textbf{(1)} retrieving the most relevant posts for each item using an adaptive dense retrieval approach; \textbf{(2)} generating responses in a zero-shot setting with LLMs, using the retrieved documents as context.

\section{Methodology} \label{meth}

\subsection{Datasets}

Our analyses uses seven datasets from the 2017, 2019, 2020, and 2022 eRisk collections \cite{losada2017erisk,losada2019overview,losada2020erisk,parapar2022erisk}, encompassing different MHCs.
The depression datasets consist of two distinct tasks. The ``\textit{early detection task dataset}'' from eRisk 2017 (used in Section \ref{depr}) provides binary classification data (depression vs. control users) through user posts and comments, with users labeled based on signs of depression. The ``\textit{severity assessment task datasets}'' from eRisk 2019 and 2020 (used in Section \ref{bdi}) contain user post histories along with their responses to the BDI-II questionnaire, enabling detailed depression severity measurement.
The self-harm, anorexia, and pathological gambling datasets belong to the \textit{``early detection tasks datasets''} from eRisk 2019, 2020, and 2022 (used in Section \ref{selfharm}). These datasets contain posts preceding users' entry into self-harm, anorexia, and gambling communities, aiming to identify early warning signals before explicit help-seeking behavior.
Table \ref{stat1} summarizes the user and post distributions across conditions for the \textit{early detection task datasets}, while Table \ref{stat2} presents detailed statistics for the \textit{severity assessment task datasets}, including the distribution of depression severity levels.

\subsection{Adaptive Zero-Shot Retrieval Strategy}

In this subsection, we address the challenge of retrieving relevant social media content for psychological assessment. The number of Reddit posts per user varies widely, leading to potential issues such as exceeding the LLM token limit and reasoning degradation due to large inputs \cite{fraga2024challenging,li2024long,li-etal-2025-enhancing-retrieval}. To mitigate these issues, we adopt a fully unsupervised retrieval strategy based on embedding similarity, exploring 10 different dense retrieval models (see Table \ref{stat_retr}) and evaluating their effectiveness through LLM prediction accuracy.
\begin{table}[t]
\footnotesize
    \centering
    \footnotesize  

    \setlength{\tabcolsep}{4pt} 
    \begin{tabular}{llcccc}
        \toprule
        Collection  & \multicolumn{2}{c}{\# of Users} &  \multicolumn{2}{c}{\# of Posts}   \\
        \cmidrule(lr){2-3} \cmidrule(lr){4-5} %\cmidrule(lr){7-8}
        &  Patient & Control & Patient  & Control   \\
        \midrule
        \scriptsize \textbf{eRisk 2017} -
        Depression & 52 & 349  & 359.7& 623.7  \\
        \scriptsize \textbf{eRisk 2019} -
         Self-Harm & 41 & 299 & 168.9 & 212.4 \\
        \scriptsize \textbf{eRisk 2020} -
        Self-Harm & 104&319  & 112.4 & 285.7  \\
       \scriptsize \textbf{eRisk 2019} - Anorexia & 73 & 742 & 241.4 & 745.1\\
       \scriptsize \textbf{eRisk 2022} - Gambling & 81 & 1998 & 180.6 & 507.6\\
        \bottomrule
    \end{tabular}
        \caption{Users %summary 
        statistics of ``early detection task datasets''  for eRisk 2017, 2019, 2020, and 2022 for \textit{depression}, \textit{self-harm}, \textit{anorexia}, and pathological gambling. }
    \label{stat1}
\end{table}
\begin{table}[t]
    \centering
    \small
    \setlength{\tabcolsep}{4pt}
    \begin{tabular}{lcccc}
        \toprule
        Collection & \textit{Minimal} & \textit{Mild} & \textit{Moderate} & \textit{Severe} \\
        \midrule
        \multirow{3}{*}{\textbf{eRisk 2019}}
         & 4 & 4 & 4 & 8 \\
        \cmidrule{2-5}
        & \multicolumn{2}{l}{\textbf{\# of Users}: 20}\\
         & \multicolumn{4}{l}{\textbf{\# of Posts}: 10'380} \\
        \midrule
         \multirow{3}{*}{\textbf{eRisk 2020}}
        & 10 & 23 & 18 & 19 \\
        \cmidrule{2-5}
                & \multicolumn{2}{l}{\textbf{\# of Users}: 70}\\
         & \multicolumn{4}{l}{\textbf{\# of Posts}: 33'600} \\
        \bottomrule
\end{tabular}
\caption{Users summary statistics of ``severity assessment task datasets''  for eRisk 2019 and 2020 editions.}
    \label{stat2}
\end{table}
\begin{table*}[t]
\tiny
\centering
\begin{tabular}{l|c c c c c c c c c c}
\textbf{Models} & \textbf{MiniLM-L6} & \textbf{MiniLM-L12} & \textbf{distilBERT-v4} & \textbf{T5} & \textbf{distilBERT-tas-b} & \textbf{all-mpnet} & \textbf{GIST} & \textbf{sf-e5} & \textbf{contriever} & \textbf{bge-large} \\
\hline
 \textbf{Cosine} &\normalsize $\newcheckmark$ & \normalsize$\newcheckmark$ &\normalsize $\newcheckmark$ &\normalsize $\newcheckmark$ & \normalsize$\newcrossmark$ &\normalsize  $\newcheckmark$ &\normalsize $\newcheckmark$ &\normalsize $\newcheckmark$ &\normalsize $\newcrossmark$ &\normalsize $\newcrossmark$\\
\hline
 \textbf{Emb. Dim.}& \normalsize384 & \normalsize384 & \normalsize768 & \normalsize768 & \normalsize768 & \normalsize768 & \normalsize768  & \normalsize1024  & \normalsize768 &\normalsize 1024\\
\hline
 \textbf{Avg. Docs retrieved} & \normalsize9 & \normalsize15  &\normalsize 9  &\normalsize 15 & \normalsize 9 & \normalsize 20  & \normalsize17 & \normalsize14& \normalsize10  &\normalsize13\\
\hline
\end{tabular}
\caption{Dense retrievers used are reported, along with how the similarity scores are calculated (\textit{cosine similarity} - $\newcheckmark$ - or \textit{dot product} - $\newcrossmark$  ), the dimension of the retriever embeddings, and the average number of documents retrieved per item with respect to the eRisk 2019 dataset.}
\label{stat_retr}
\end{table*}
To account for variability in relevant posts per user, we employ the ABIDE-ZS method \cite{ravenda2024}. 
For each item%-query (\(iq\))
, this approach identifies a neighborhood where the semantic meaning remains stable - i.e., the posts within this region share contextual relevance to the specific questionnaire's item.
Posts are retrieved based on semantic similarity, selecting the top \(k^*\) posts, where \(k^*\) is optimally determined by the ABIDE algorithm \cite{dinoia2024} for each item (see Section \ref{abide_app}). This eliminates the need to fix a priori the number of $k$ posts to retrieve or set a threshold, making our approach \textit{adaptive}.

Figure \ref{steps} illustrates the retrieval process. For a each user \(i\), embeddings are generated for their posts (scatters) and for the four response choices (denoted by the \(\star\) symbol) within an item. Let \(\mathbf{P}_i = \{\mathbf{p}_{i1},  \ldots, \mathbf{p}_{im}\}\) represent the embeddings of the $m$ posts for user \(i\), and \(\mathbf{iq}_j = \{\mathbf{iq}_{j1}, \mathbf{iq}_{j2}, \mathbf{iq}_{j3}, \mathbf{iq}_{j4}\}\) represent the embeddings of the four 
choices for item \(j\). These item choices serve as queries (aka item-queries) to retrieve the most relevant posts from the user's history. 
For each choice \(\mathbf{iq}_{jl}\) (where \(l = 1, 2, 3, 4\)), the relevant Reddit posts \(\mathbf{RRP}_{jl}\) (represented by colored dots in Figure \ref{steps}) are retrieved by selecting the top \(k^*\) posts with the highest embedding similarity scores:
\(
\mathbf{RRP}_{ijl} = \{p_{i}\}_{p_i \in \aleph(iq_{jl})}
\),
where \(\aleph(IC_{jl})\) represents the adaptive neighborhood of size \(k^*\) for item \(j\) and choice \(l\).
Consider user posts and questionnaire item choices  embedded in ${R}^D$, where $D$ is the dimension of the embeddings. The optimal \(k^*\) is determined by the local density of posts around each query point, ensuring we retrieve exactly the number of posts needed to maintain semantic coherence. 
In other words, the space identified by the adaptive neighborhood for each item-query can be seen as a space where the semantics of the item-query remain constant, which is reflected in the region where the posts are semantically relevant to the item-query.
Table \ref{stat_retr} shows the statistics of the different dense retrieval models used, including the similarity measure adopted for each retriever (a more in-depth discussion about retrievers used can be found in Appendix \ref{bench}). We observe that while the number of documents retrieved per item varies across models, no significant correlation exists between embedding dimension and the average number of retrieved documents (Pearson correlation, \(p-value = 0.70\)). The impact of $k^*$ is discussed in Section 
\ref{adaptkstars} of the Appendix.

\begin{figure}
    \centering
    \includegraphics[width=1.\linewidth]{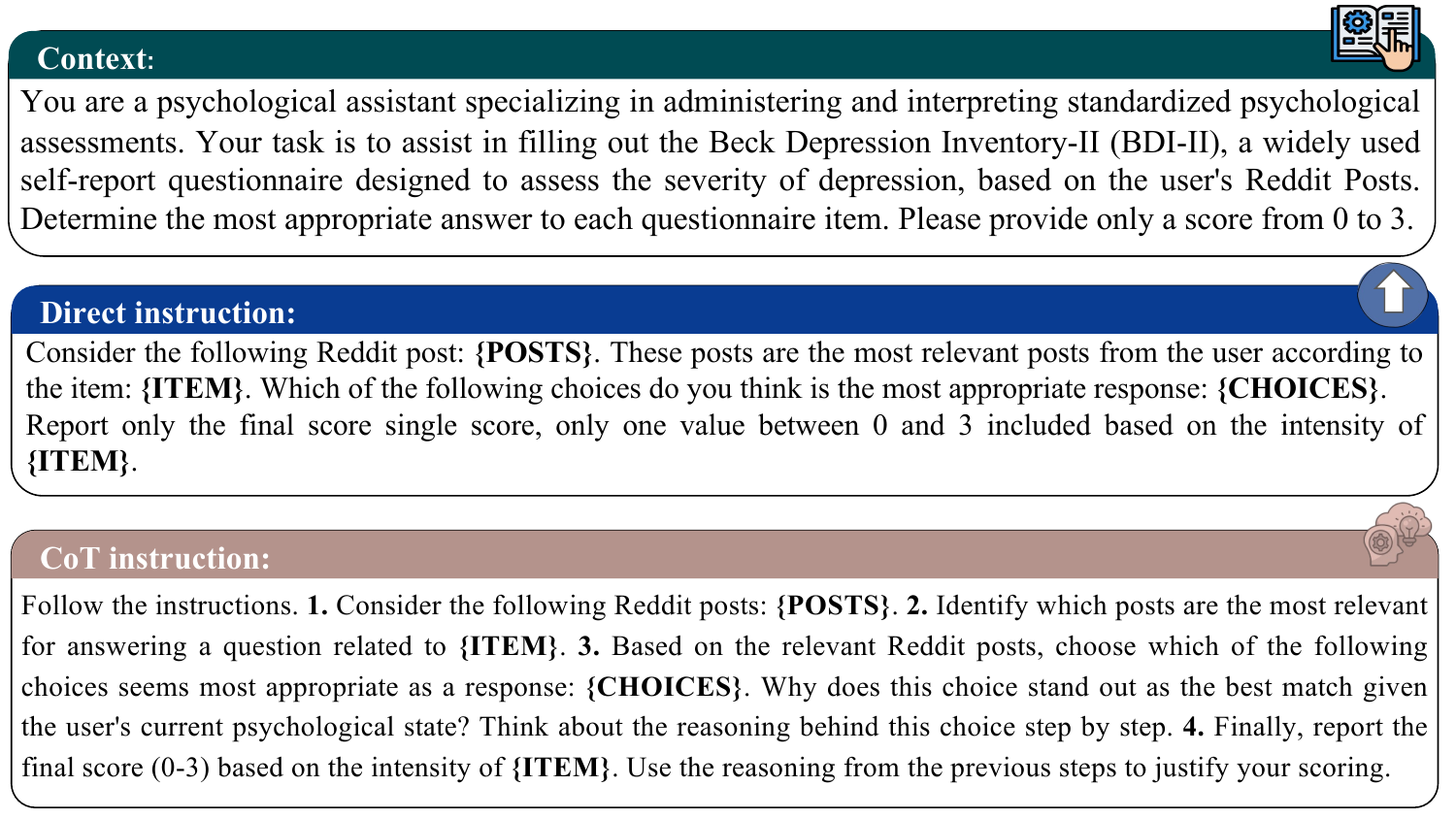}
    \caption{Prompt templates based on different prompting strategies: Direct and CoT.}
    \label{prompts}
\end{figure}
\subsection{Proposed Framework}

The proposed workflow - aRAG - involves a two-step pipeline. First, we retrieve the most relevant posts for each user with respect to each questionnaire item, and then we instruct the LLM to predict the corresponding score.
\begin{figure*}[ht]
    \centering
    \includegraphics[width=.95\linewidth]{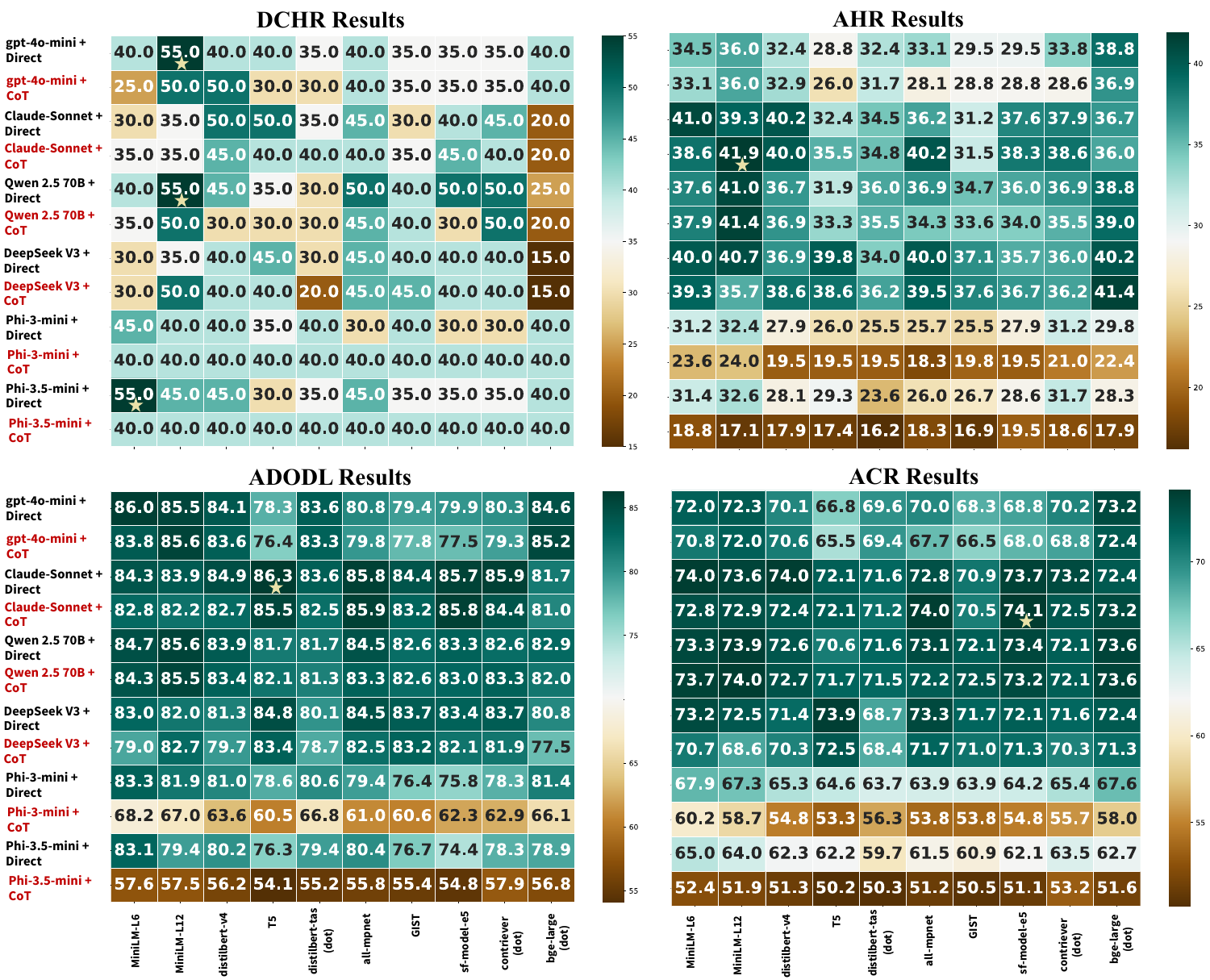}
    \caption{Results associated with the use of different combinations of LLMs, prompting strategies, and retrieval models across various metrics for the eRisk 2019 collection. The best combinations for each metric are highlighted with $\star$. The greener the color, the better the score. For readability we report only the first decimal.}
    \label{heat}
\end{figure*}
Our investigation  combines SOTA retrieval techniques with both open and closed-source LLMs, examining how different prompting strategies affect psychological assessment accuracy. All LLMs are used with temperature set to 0 to force deterministic outputs.
For predictions, we use the closed-source models \texttt{gpt-4o-mini} and \texttt{Claude-3.5-Sonnet}, alongside large-scale open-source models \texttt{Qwen 2.5 70B} and \texttt{DeepSeek V3}, as well as lightweight open-source models \texttt{Phi-3-mini} and \texttt{Phi-3.5-mini}.
For prompting, we use two techniques illustrated in Figure \ref{prompts}. The first technique predict the item scores directly based on relevant Reddit posts (Direct), while the second approach guides the LLM to reflect on intermediate steps (CoT), encouraging the LLM to go through reasoning steps before predicting the final score.

\section{Results} \label{res}

\begin{table*}[htbp]
    \centering
    \small  
    \setlength{\tabcolsep}{4pt} 
    \begin{tabular}{lllcccc}
        \toprule
        Collection & Model + Prompt Strategy &  Retrieval &\multicolumn{2}{c}{Questionnaire Metrics} &  \multicolumn{2}{c}{Item Metrics} \\
        \cmidrule(lr){1-5} \cmidrule(lr){6-7}
        & & &DCHR  & ADODL  &  AHR  & ACR \\
        \midrule
        \multirow{7}{*}{\rotatebox[origin=c]{90}{eRisk 2019}} 
        & CAMH  & & 45.00\%  & 81.03\% & 23.81\%   & 57.06\%  \\
        & UNSLC \cite{burdisso2019unsl} &  & 40.00\%& 78.02\%&41.43\%  &69.13\%  \\
        & UNSLE \cite{burdisso2019unsl}&  & 35.00\%& 80.48\% & 40.71\%   & 71.27\% \\
        \cdashline{2-7}[1pt/1pt]
        &  \texttt{Qwen 2.5 70B} + \textbf{Direct}  &  MiniLM-L12 & \textbf{55.00}\% & 85.56\%  & 40.95\%  & 73.81\%  \\
        & \texttt{gpt-4o-mini} + \textbf{Direct}  &  MiniLM-L12& \textbf{55.00}\% & 85.48\%  & 35.95\% & 72.30\% \\
        &  \texttt{Claude Sonnet} +  \textbf{Direct}  &  T5  & 50.00\%  & \textbf{86.27}\%    & 32.38\% & 72.14\%  \\
        &   \texttt{Claude Sonnet} +  \textbf{CoT}    & MiniLM-L12 & 35.00\%& 82.22\% & \textbf{41.90}\%   & 72.86\% \\
        &   \texttt{Claude Sonnet} +  \textbf{CoT}    &  sf-model-e5& 45.00\%&  85.79\% & 38.33\%  & \textbf{74.12}\% \\
        \midrule
        \multirow{11}{*}{\rotatebox[origin=c]{90}{eRisk 2020}} 
        & ILab \cite{martinez2020early} &  &27.14\% & 81.70\% &  37.07\% & 69.41\% \\
        & Relai \cite{maupome2020early} &  &34.29\% & 83.15\% & 36.39\% & 68.32\% \\
        \cmidrule[1pt](lr){2-7}
        & Sense2vec \cite{perez2022automatic} &  & 37.14\% & 82.61\%  & 38.97\% & 70.10\% \\
        & \cite{perez2023bdi} Recall & &50.00\%  & 85.24\% & 35.44\% & 67.23\% \\
        & \cite{perez2023bdi} Voting & &47.14\%  & \textbf{85.33}\% & 35.24\% & 67.41\% \\
        \cdashline{2-7}[1pt/1pt]
        & \texttt{Qwen 2.5 70B} + \textbf{Direct}  &  MiniLM-L12 & 41.43\% & 83.49\%  & 38.78\%  & 72.74\%  \\
        & \texttt{gpt-4o-mini} + \textbf{Direct}  &  MiniLM-L12 & 41.43\% & 84.01\%  & 36.60\%  & 71.59\%  \\
        & \texttt{Claude Sonnet} +  \textbf{Direct}  & T5  & 47.14\% & 83.92\% & 39.52\% & 73.31\% \\
        &   \texttt{Claude Sonnet} +  \textbf{CoT}    & MiniLM-L12 &32.86\%& 81.59\% & \textbf{41.90}\%   & 72.56\% \\
        &  \texttt{Claude Sonnet} +  \textbf{CoT}  & sf-model-e5 & 42.86\%  & 84.17\% & 41.77\% & 73.83 \% \\
        & \texttt{LLMs Ensemble} & - & \textbf{52.86}\% & 84.63\% & 39.52\% & \textbf{74.10}\%   \\
        \bottomrule
    \end{tabular}
        \caption{Model performance comparison on eRisk 2019 and 2020 collection w.r.t. questionnaire metrics (DCHR, ADODL) and item metrics (AHR, ACR). Bold values represent the best results for each collection. For all the considered metrics, the higher the score, the better.}
    \label{model_comparison}
\end{table*}

\begin{table}[htbp]
    \centering
    \tiny  

    \setlength{\tabcolsep}{4pt}  
    \begin{tabular}{lllcccc}
        \toprule
        Collection & Model + Prompt Strategy &  Retrieval & DCHR\ BDI-II &  DCHR BDI \\
        \midrule
        \multirow{4}{*}{\rotatebox[origin=c]{90}{eRisk 2019}} 
        & \texttt{gpt-4o-mini} + \textbf{Direct}  &  MiniLM-L12& 55.00\% & 55.00\% \\
        &  \texttt{Claude Sonnet} +  \textbf{Direct}  &  T5  & 45.00\%&50.00\%  \\
        &   \texttt{Claude Sonnet} +  \textbf{CoT}    & MiniLM-L12 &\textbf{50.00\%} &35.00\%\\
        &   \texttt{Claude Sonnet} +  \textbf{CoT}    &  sf-model-e5& \textbf{55.00\%}& 45.00\% \\
        \midrule
        \multirow{4}{*}{\rotatebox[origin=c]{90}{eRisk 2020}} 
        & \texttt{gpt-4o-mini} + \textbf{Direct}  &  MiniLM-L12 & \textbf{42.86\%} &41.43\%  \\
        & \texttt{Claude Sonnet} +  \textbf{Direct}  & T5  & \textbf{48.57\%}&47.14\% \\
        &   \texttt{Claude Sonnet} +  \textbf{CoT}    & MiniLM-L12 &  \textbf{50.00\%} &32.86\%\\
        &  \texttt{Claude Sonnet} +  \textbf{CoT}  & sf-model-e5 &\textbf{54.29\%} &42.86\% \\
        \bottomrule
    \end{tabular}
        \caption{Performance comparison regarding the correct categorization of depressive state intensity considering the BDI-II true reparametrization.}
    \label{dchr}
\end{table}

\subsection{Predicting Psychological Questionnaire Scores} \label{bdi}

To evaluate the effectiveness of our approach in predicting responses to the BDI-II questionnaire, we use the official eRisk benchmark metrics \cite{losada2019overview} that assess performance at two distinct levels (for all the metrics considered, the higher the value, the better):  

At \textit{the level of the questionnaire}, we examine the Hit Rate of the Depression Category (DCHR) which measures the accuracy in estimating depression severity levels (minimal: \textbf{0-9}, mild: \textbf{10-18}, moderate: \textbf{19-29}, and severe: \textbf{30-63}), and the Average Difference between Overall Depression Levels (ADODL), which evaluates the general BDI-II score estimations. 

At \textit{item level}, we employ Average Hit Rate (AHR) to evaluate prediction accuracy for individual symptoms, and Average Closeness Rate (ACR) to measure how close predictions are to actual values for each symptom.

We use datasets from the 2019 and 2020 eRisk editions, which contain users' post histories and their responses to the BDI-II questionnaire.

We systematically evaluate different combinations of LLMs, dense retrievers, and prompting strategies for the eRisk 2019 dataset, using the best combinations for the eRisk 2020 collection. 
To address \textbf{(RQ.2)}, Figure \ref{heat} presents heatmap visualizations comparing performance across all combinations for the four evaluation metrics on the eRisk 2019 dataset. 

On average, we observe that lightweight models perform significantly better with direct prompting compared to CoT approach, while this difference is less pronounced in larger architectures. Overall, we notice that closed-source LLMs and large-scale open-source models often outperform lightweight ones. This pattern can be attributed to the limited model capacity of lightweight architectures, their reduced ability to effectively manage multi-step reasoning chains during CoT prompting, and their lesser capability to capture subtle linguistic nuances crucial for mental health assessment. Specifically, the worst results are obtained when using lightweight open-source models combined with CoT prompting techniques.

After identifying the best performing combinations on the eRisk 2019 dataset from Figure \ref{heat}, we use these to the 2020 collection. Table \ref{model_comparison} shows these results alongside baseline benchmarks.
For comparison, we consider the best performing models from previous work for each metric for the eRisk 2019 and 2020 collections. We refer the reader to the corresponding overview for more in-depth details \cite{losada2019overview,losada2020erisk}.  In Section \ref{bench} of the Appendix we further discuss all the models used as benchmarks. 
Regarding the 2019 edition, the proposed approaches outperform the benchmarks across all considered metrics, except for AHR, where only one combination manages to outperform that edition's best model. 
On the 2020 eRisk collection, we achieve:
(1) superior item-level metrics; our approach outperforms existing benchmarks on granular metrics (AHR and ACR); (2)  SOTA depression category accuracy (DCHR); we obtain the best result with a voting-regressor ensemble based on the rounded average scores of the top 3 closed-source approaches with the highest DCHR scores in 2019 evaluations (as reported in Table \ref{model_comparison}).
These results are particularly notable as we maintain high performance without requiring training data, unlike all previous approaches reported that rely on the 2019 dataset for supervision.
Interestingly, as shown in Table \ref{model_comparison}, our approach demonstrates consistent performance across both eRisk editions despite their different category distributions (see Table \ref{stat2}). Specifically, our method maintains stable performance on three key metrics (ADODL, AHR, and ACR), with only DCHR showing variation between editions. % , where the difference is not statistically significant when considering the result vectors of the 4 methods. 
This negative result is due to how overall scores are categorized into the 4 severity categories of the BDI-II questionnaire. Within the context of the challenge, the authors of the eRisk workshop use the BDI parameterization, the version preceding BDI-II. In BDI-II, new ranges are introduced that change from those of the previous test, especially regarding the minimum level of depression. Specifically, minimal or absent depression is identified as \textbf{0-13}, Mild as \textbf{14-19}, Moderate as \textbf{20-28}, and severe as \textbf{29-63} \cite{beck1996beck,warmenhoven2012beck}. Table \ref{dchr} shows how DCHR changes when using the correct reparameterization, obtaining excellent and completely counterintuitive results, especially for the two models using Claude and the CoT strategy, compared to those obtained with the previous questionnaire parameterization in the 2020 dataset.
In Section \ref{nonretr} of the Appendix, we further justify our aRAG approach's effectiveness by comparing it against non-RAG baselines, where we test closed-source LLMs (\texttt{gpt-4o-mini} and \texttt{Claude-3.5-Sonnet}) using direct input of all posts within their context window. 

\begin{figure*}
    \centering
    \includegraphics[width=.83\linewidth]{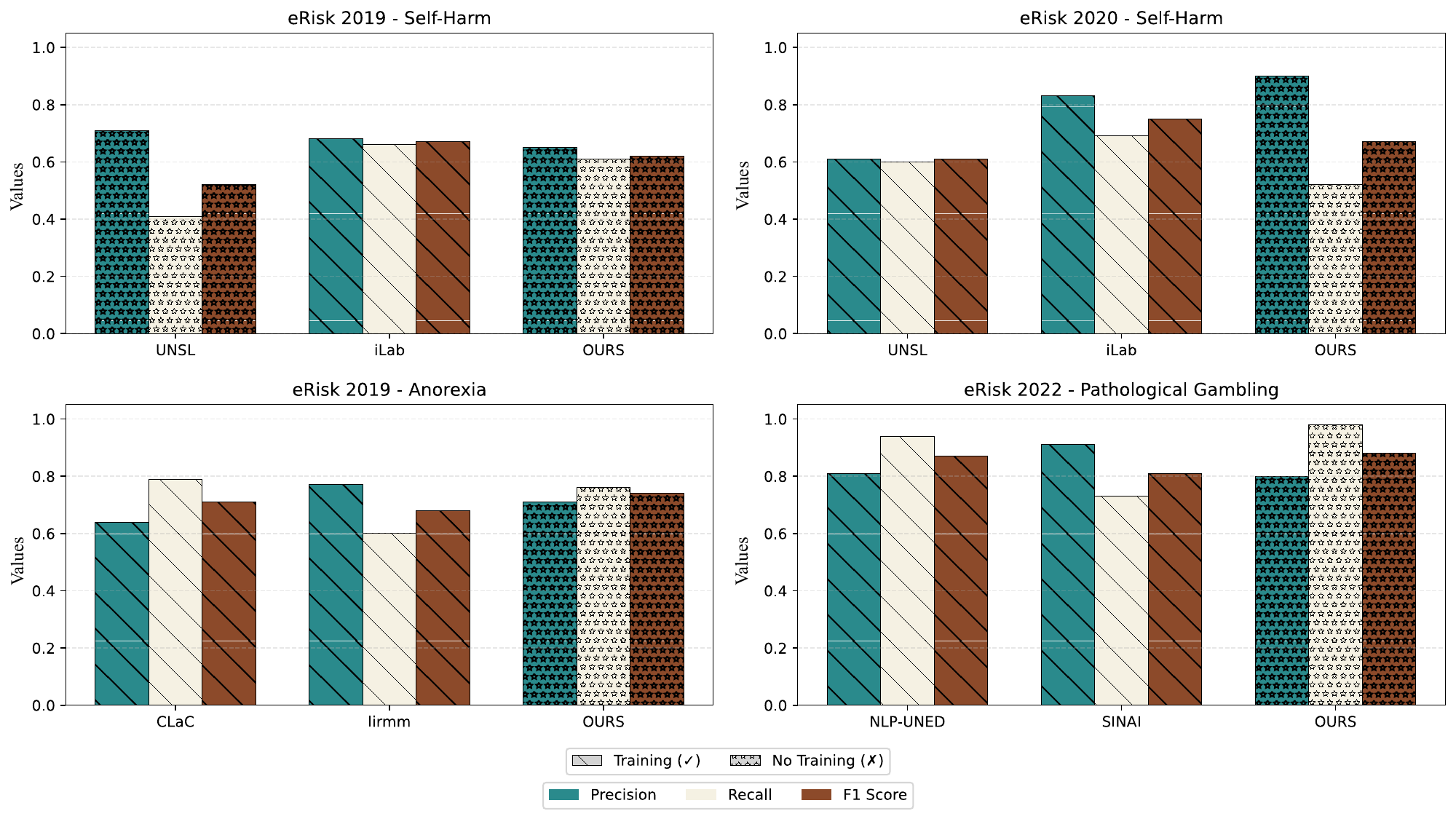}
    \caption{Performance of our approach compared to the best approach in each of the two eRisk editions for the  detection of self-harm, Anorexia, and Pathological Gambling. Models with $\newcheckmark$ use training data while models with $\newcrossmark$ are fully unsupervised. Precision, Recall, and F1 metrics are reported.}
    \label{self_tab1}
\end{figure*}

\subsection{Beyond Depression: Identifying Signs of different MHCs through questionnaire } \label{selfharm}

In this section, as a proof-of-concept, we extend the use of questionnaires to different mental health conditions, such as self-harm, anorexia, and pathological gambling.
We approach the identification of self-harm behaviors using the Self-Harm Inventory (SHI), a 22-item, yes/no self-report questionnaire, that screens for lifetime history of self-harm behaviors. A score of 5 or more ``yes'' responses on the SHI indicates potential mild forms of Deliberate Self-Harm (DSH)~\cite{latimer2009psychometric}. 
To answer \textbf{(RQ.3)}, the methodology follows the same approach used for BDI-II, with the key distinction that SHI is structured as a binary questionnaire, unlike BDI-II Likert scale. The process involves retrieving the most relevant posts using SHI questions as queries, followed by LLM-generated responses. For SHI, since each item has binary yes/no responses, we use only the questions as retrieval queries.
For this specific task, we chose the combination of gpt-4o-mini with Direct Prompt and MiniLM-L12-v3 as the retrieval model, as this configuration demonstrated the best trade-off between performance, computational costs, and processing speed in previous tests.
The dataset provides each user complete post history, associated with a binary label indicating the presence or absence of self-harm behaviors. While the original task aims to identify self-harm cases as early as possible using the minimum number of posts, our approach considers the entire history to maximize prediction accuracy.
For the 2019 edition, we compared our approach with two reference models: UNSL \cite{burdisso2019unsl}, which analyzes only a subset of posts, and iLab \cite{martinez2020early}, which uses BERT fine-tuned in a fully supervised context. iLab approach was optimized to maximize F1-score and trained on a custom dataset built from self-harm subreddit posts. To ensure a fair comparison, we considered the version of iLab that has access to the complete post history for both the 2019 and 2020 editions.
Figure \ref{self_tab1} shows that while the fine-tuned BERT-based model, benefiting from an extensive training corpus, generally outperforms our unsupervised approach, the performance gap is remarkably narrow, particularly for the 2019 edition. Notably, our approach achieves superior precision in the 2020 edition. %
These results are particularly significant considering that our approach do not need any training data and requires no training data, yet achieves competitive performance compared to fully supervised models that leverage extensive domain-specific training. 

We further applied the same methodology to the early detection of anorexia and pathological gambling, using the SCOFF questionnaire and the Pathological Gambling Diagnostic Form respectively. In both cases, we used the same pipeline as in the SHI scenario. As reported in Figure \ref{self_tab1}, our method achieves the highest F1 score for anorexia and gambling tasks, outperforming the top-scoring systems of the respective editions \cite{mohammadi2019quick,ragheb2019attentive,marmol2022sinai,fabregat2022uned} in this specific metric. These findings reinforce the generalizability of our adaptive RAG framework, demonstrating that standardized psychological questionnaires can effectively guide LLMs to detect a wide range of MHCs in a fully unsupervised setting.
This suggests that our unsupervised approach based on adaptive RAG can offer a viable alternative in scenarios where labeled training data is scarce or unavailable.

Even though in some cases the results do not outperform the benchmark models, this may be attributed not only to the lack of training data in our approach, but also to the fact that our questionnaire-guided approach occasionally fails to find sufficient evidence to support a diagnosis based on the psychometric tool used. As a result, our approach tends to be more conservative in assigning final diagnoses, sometimes refraining from matching the diagnostic criteria defined by the questionnaires unless the retrieved content provides clear and consistent indicators.

\subsection{The Importance of Questionnaire for Screening Procedure} \label{depr}

\begin{table}
\small
    \centering
    \begin{tabular}{llcccc}
        \toprule
         & Model & $\mathbf{\tau}$ & \textbf{P} & \textbf{R} & \textbf{F1} \\
        \midrule
        \multirow{3}{*}{\rotatebox[origin=c]{90}{\textbf{eRisk 2017}}} 
        & \texttt{gpt-4o-mini} & - &0.32 & 0.74 & 0.45\\
        & PHQ-9  & 10 &\textbf{0.69} & 0.45  & 0.55\\
        &  DASS (subscale) & 14 &0.36 &  0.64& 0.46 \\
        & BDI-II & 20 & 0.42  &\textbf{0.76}& 0.54 \\
        & Quest. Ensemble & & 0.53 & 0.69 & \textbf{0.60} \\
        \bottomrule
    \end{tabular}
    \caption{Performance comparison of our approach 
    %(using adaptive RAG with gpt-4-mini + Direct Prompt and MiniLM-L12-v3 for retrieval) 
    applied to different depression screening questionnaires (BDI-II, DASS-42, PHQ-9) and 
    %using the same RAG approach by
    instructing \texttt{gpt-4o-mini} to only classify users into %diagnose for the 
    depressed/non-depressed categories.}
    \label{self_tab2}
\end{table}
For this task, we used data from the 2017 eRisk edition to address \textbf{(RQ4.)}. As in the previous subsection, we have access to users post histories along with corresponding labels indicating whether each user exhibited clinical signs of depression.
To prevent trivial classification by the LLM, we removed the word depression and related terms from posts in which users explicitly self-diagnosed.

We tested the same aRAG approach that proved effective for previous tasks, using the same combination of methods but guided by different types of standardized depression screening questionnaires. Specifically, we compared results obtained from the Beck Depression Inventory-II (BDI-II), Patient Health Questionnaire-9 (PHQ-9), Depression Anxiety Stress Scales-42 (DASS-42) (specifically focusing on its depression subscale), and simply instructing the LLM to determine if the patient was suffering from depression.
While for BDI-II we used choice options as queries, for PHQ-9 and DASS-42 we only used the questionnaire items as queries since only the questions contain textual content.

Each of the three questionnaires includes an established optimal cut-off score, denoted as $\tau$ in Table \ref{self_tab2}, for identifying clinically significant depression, specifically at the moderate severity threshold. The table also reports the results obtained from the different approaches. The worst results were obtained through direct instruction to \texttt{gpt-4o-mini} and through the use of the DASS-42 depression subscale. The best results in terms of precision were achieved using PHQ-9 (0.69), while BDI-II showed the highest recall (0.76). The highest F1 score (0.60) was achieved using an ensemble classifier combining all three questionnaires.
To answer \textbf{(RQ4.)}, our findings suggest that structured psychological assessment tools, can enhance the effectiveness of LLM-based mental health assessments compared to direct questioning approaches.

\section{Conclusions} \label{conc}

We introduce a novel aRAG approach that leverages standardized psychological questionnaires to guide LLMs in mental health screening, requiring no training data.
We demonstrate the advantage of our approach in supporting mental health screening tasks. The results show the advantages of our approach in: automatically completing questionnaires \textbf{(RQ1.)}, proving effective not only for depression screening but also extending successfully to other conditions like self-harm detection, anorexia, and pathological gambling, as well as potentially other MHCs \textbf{(RQ3.)}, across different combinations of LLMs, prompts, and retrievers \textbf{(RQ2.)}.
We also show how our approach improves upon simpler methods for screening procedure that rely on direct prompting about the presence or absence of a depressive disorder \textbf{(RQ4.)}, by providing a structured interpretation of the user's psychological state and enabling an estimation of the severity level through clinically questionnaire scores.

\section{Limitations}

The proposed methodology offers several advantages in terms of its implementation and performance. Despite these, it is important to address the limitations of this approach.

Although the results are particularly promising given the available data, a limitation of this work is the relatively small number of users.
Furthermore, although the method can be easily extended to other types of questionnaires, there is no guarantee that similar results will be replicated across different questionnaires or various types of MHCs.

Additionally, while the BDI-II is considered one of the most reliable tool for depression assessment, it has some limitations. As with all self-report measures, it can be influenced by the patient subjectivity and should not replace a comprehensive clinical diagnosis. Instead, it should be used as a screening tool in conjunction with other clinical evaluation methods for a complete and accurate diagnosis.

Furthermore, in this work we use cut-off scores to define different risk thresholds for specific disorders as reported in the original works based on psychometric criteria. However, these cut-off score guidelines are typically provided with the recommendation that thresholds should be adjusted according to sample characteristics and the intended purpose of the questionnaire. Additionally, such cut-offs may not be fully consistent in the context of social media analysis and may require further adaptation.

\section{Ethical Considerations}

The proposed methodology for mental health support and assessment, while novel, raises several ethical considerations that must be addressed to ensure responsible deployment. 

There is potential for AI to be misused as a clinical tool. Without proper safeguards, these models could exhibit harmful or biased behaviors. 
It is crucial to emphasize that this approach should not be viewed as a substitute for specialized medical professionals, but rather as a method to screen for potential subjects at risk of depression.

Ethical considerations extend to privacy and data protection, ensuring the confidentiality and security of users social media data. 
\\

\noindent \textbf{Acknowledgements.} This research has partly been founded by the SNSF (Swiss National Science Foundation) grant 200557.

\bibliography{custom}

\appendix

\newpage

\begin{figure*}[ht]
    \centering
    \includegraphics[width=1\linewidth]{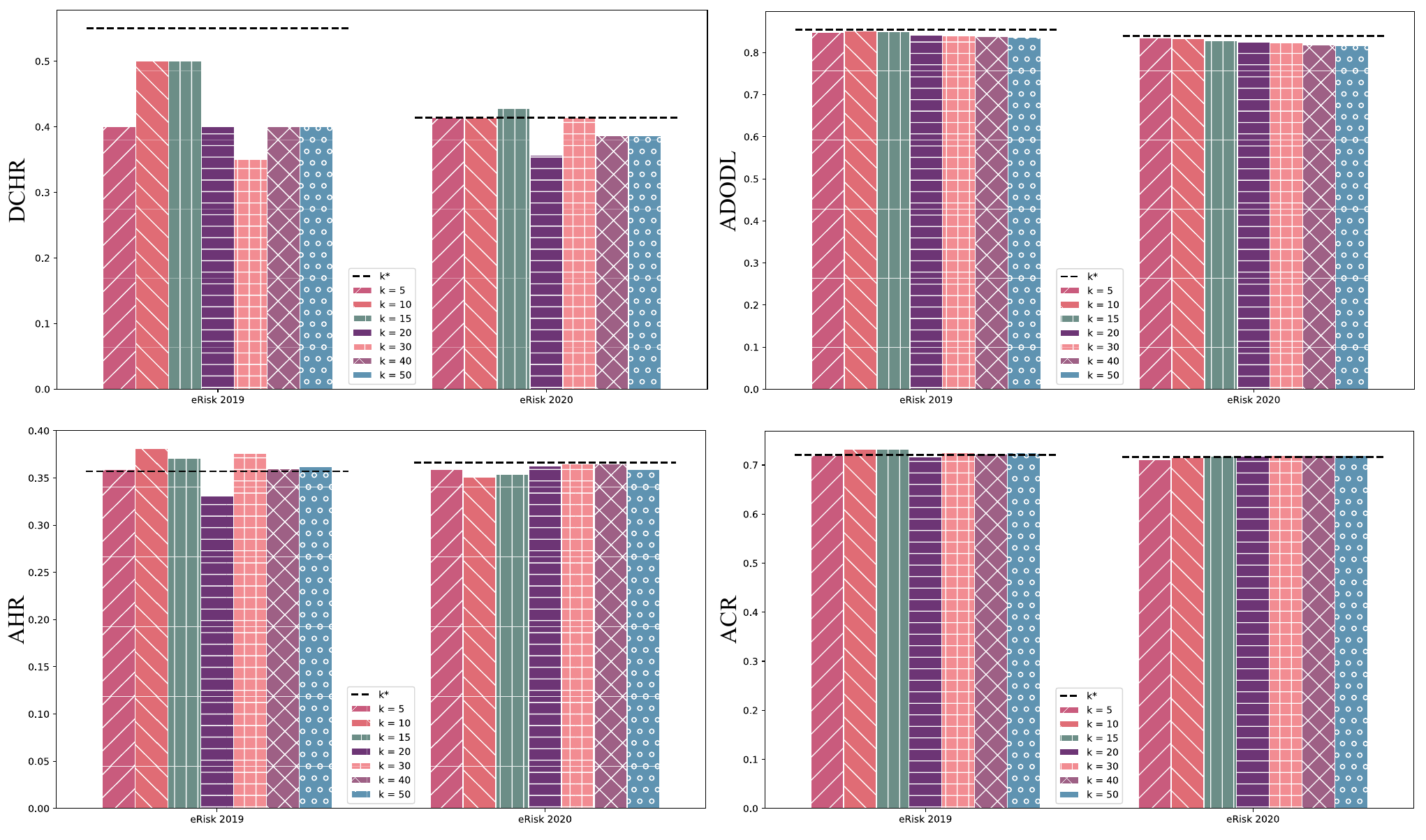}
    \caption{Comparison of the performance of the gpt-4o-mini + Direct Prompt combination with MiniLM-L12-v3 retrieval model on the eRisk 2019 and 2020 datasets, varying by $k$ value used to select the number of documents to retrieve. $k^*$ represent our adaptive method, and $k=15$ the average number of $k^*$.}
    \label{kstars}
\end{figure*}

\section{Data Availability}

The datasets supporting the conclusions of this article from eRisk collections are available for research purposes under signing user agreements.

\section{ABIDE} \label{abide_app}

\textit{Adaptive Binomial Intrinsic Dimension Estimator} (ABIDE) \cite{dinoia2024} is an approach to estimate the Intrinsic Dimension (ID) of the data. The idea of ID is to quantify the complexity of high-dimensional datasets. In essence, it represents the minimum number of variables needed to describe the underlying structure of data without significant loss of information \cite{denti2022generalized,denti2023bayesian}.
In fact, in many real-world scenarios, especially for what concerns text, there are often hidden relationships and dependencies among these features. This means that the data might actually lie on a lower-dimensional manifold embedded within the high-dimensional space.

However, estimating ID can be difficult, especially in real-world scenarios, where datasets often have complex structures and ID can vary depending on the scale at which the data are observed.

ABIDE addresses the scale-dependency challenge through a novel adaptive approach. At its core, ABIDE uses the concept of $k^*$, which represents the optimal neighborhood size for each data point. $k^*$ is not fixed across all the observations, but varies for each point, allowing the method to adapt to local data characteristics.
The algorithm works iteratively. It starts with an initial ID estimate, using the 2NN method \cite{facco2017estimating}. Then, for each data point, it determines the largest neighborhood ($k^*$) where the data density can be considered approximately constant. 
Using these $k^*$ values, ABIDE recalculates the ID estimate. This process is repeated, refining at each iteration both the ID estimate and the $k^*$ values for each point. The iteration continues until convergence is reached.

\section{Dense Retrieval Models}

We use a  pool of different dense retrieval models: \textit{msmarco-MiniLM-L-6-v3\footnote{\href{https://huggingface.co/sentence-transformers/msmarco-MiniLM-L-6-v3}{https://huggingface.co/sentence-transformers/msmarco-MiniLM-L-6-v3}}, msmarco-MiniLM-L-12-v3\footnote{\href{https://huggingface.co/sentence-transformers/msmarco-MiniLM-L-12-v3}{https://huggingface.co/sentence-transformers/msmarco-MiniLM-L-12-v3}}, msmarco-distilbert-base-v4\footnote{\href{https://huggingface.co/sentence-transformers/msmarco-distilbert-base-v4}{https://huggingface.co/sentence-transformers/msmarco-distilbert-base-v4}}, sentence-T5-base\footnote{\href{https://huggingface.co/sentence-transformers/sentence-t5-base}{https://huggingface.co/sentence-transformers/sentence-t5-base}}, msmarco-distilbert-base-tas-b\footnote{\href{https://huggingface.co/sentence-transformers/msmarco-distilbert-base-tas-b}{https://huggingface.co/sentence-transformers/msmarco-distilbert-base-tas-b}}, all-mpnet-base-v2\footnote{\href{https://huggingface.co/sentence-transformers/all-mpnet-base-v2}{https://huggingface.co/sentence-transformers/all-mpnet-base-v2}}, GIST\footnote{\href{https://huggingface.co/avsolatorio/GIST-Embedding-v0}{https://huggingface.co/avsolatorio/GIST-Embedding-v0}}, sf-model-e5\footnote{\href{https://huggingface.co/jamesgpt1/sf_model_e5}{https://huggingface.co/jamesgpt1/sf\_model\_e5}}, contriever-msmarco\footnote{\href{https://huggingface.co/facebook/contriever-msmarco}{https://huggingface.co/facebook/contriever-msmarco}}, bge-large\footnote{\href{https://huggingface.co/BAAI/bge-large-en}{https://huggingface.co/BAAI/bge-large-en}}}.
The selected models are a diverse and representative sample of the SOTA for dense retrieval and have already been tested in the literature \cite{khramtsova2023selecting,barros2024anchor,solatorio2024gistembed}.

The chosen models, mostly pre-trained on MS MARCO \cite{bajaj2018msmarcohumangenerated}, allow for an in-depth analysis of generalization capabilities in zero-shot scenarios. From an empirical validation perspective, many of the selected models are present in the BEIR leaderboard \cite{thakur2021beirheterogenousbenchmarkzeroshot}, thus providing established benchmarks for performance evaluation. The standardization of implementation through the Hugging Face platform ensures uniformity in evaluations and facilitates the reproducibility of results.

\section{Benchmark Models} \label{bench}

As benchmark models for ``\textit{Measuring the Severity of depression}'' task, we use a combination of top-performing models from the eRisk competition and SOTA models from the literature on the considered datasets, specifically regarding the task of measuring depression severity.

For eRisk 2019, two notable approaches were developed. The first is CAMH \cite{losada2019overview}, which represented users through LIWC features. Then, for each BDI questionnaire item, it matches a vectorial representation of the user against vectorial representations of possible responses. The second approach, UNSL \cite{burdisso2019unsl}, converts textual indicators from user posts into a standardized clinical depression score (0-63). It maps linguistic analysis into 4 clinical severity categories using various statistical and text processing techniques to complete the 21-question diagnostic questionnaire.

For eRisk 2020, several methods emerges. BioInfo \cite{oliveira2020bioinfo} and Relai \cite{maupome2020early} methods obtained their own datasets to perform standard ML classifiers using engineered features as linguistic markers. 

We also refer to recent works \cite{perez2022semantic,perez2023bdi}. The two approaches aim to estimate depression severity from Reddit posts using BDI-II symptom-based classifiers. While the first approach \cite{perez2022semantic} uses word embeddings to compare BDI-II options and user texts, \cite{perez2023bdi} leverages expert-annotated ``golden'' sentences (738 in total) as queries to identify semantically similar ``silver'' sentences through RoBERTa embeddings, achieving better performance through Accumulative Voting and Recall aggregation methods.

\section{Ablation Study}

\subsection{Justification of the RAG approach}\label{nonretr}

\begin{figure*}
    \centering
    \includegraphics[width=1\linewidth]{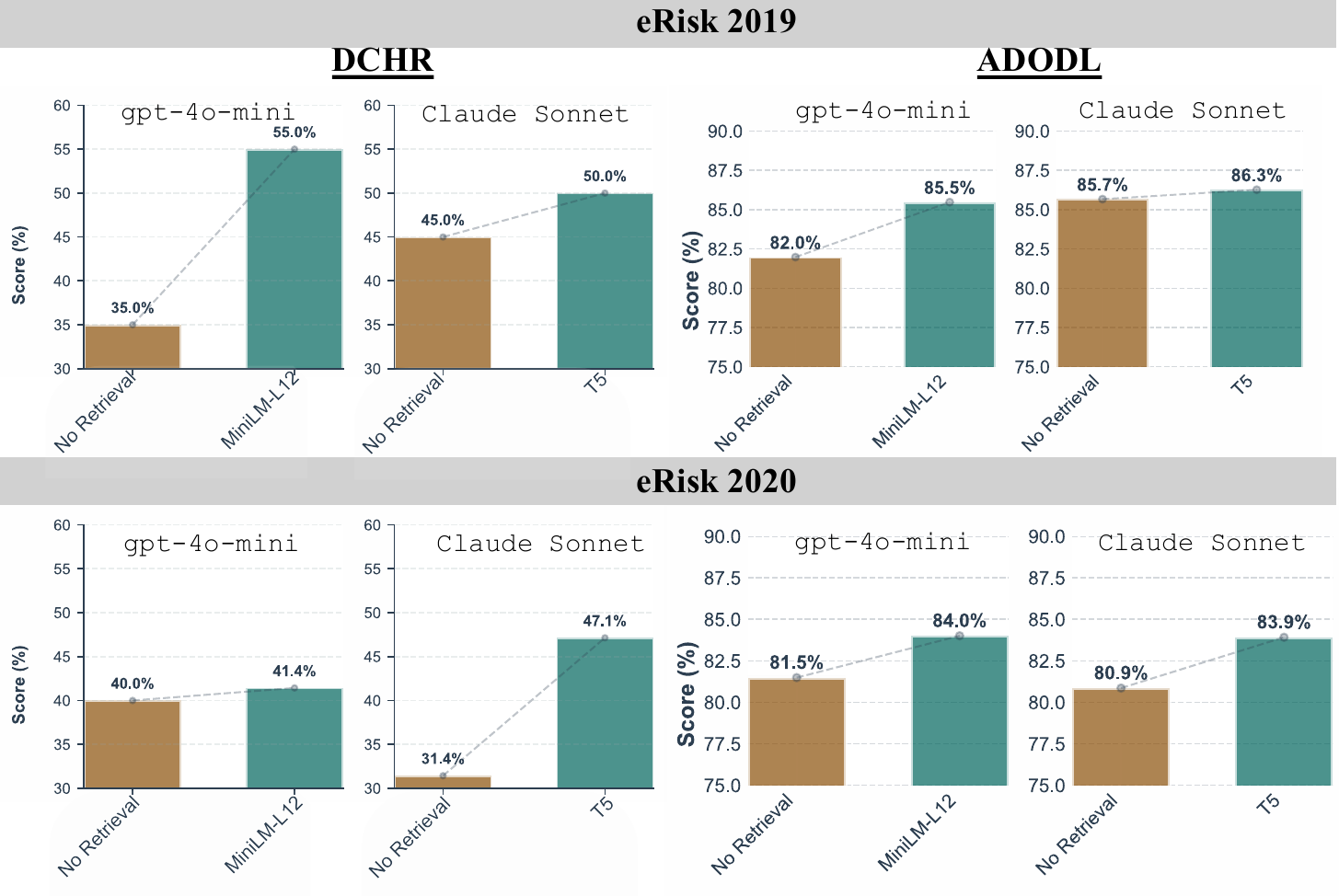}
    \caption{Performance comparison between aRAG and no-retrieval approaches for closed-source LLMs across DCHR and ADODL metrics on eRisk 2019 and 2020 datasets}
    \label{retrnoretr}
\end{figure*}

To further justify our retrieval-based approach, aka aRAG, we compare the performance of the most effective DCHR configurations using two closed-source LLMs against their performance when directly prompted to answer BDI-II items without filtering relevant posts. Specifically, we compare Claude 3.5 Sonnet + Direct prompt combined with T5 and gpt-4o-mini + Direct prompt combined with MiniLM-L12-v3. Given that some users have a significantly high number of posts, we input all posts that fit within each LLM's context window based on timestamp order. We focus on the questionnaire-level metrics, DCHR and ADODL, which allow us to assess the accuracy of both approaches in determining depression severity. As shown in Figure \ref{retrnoretr}, the aRAG approach consistently outperforms its no-retrieval counterpart across both collections and LLMs. Notably, we observe particularly wide performance gaps when using gpt-4o-mini on the eRisk 2019 collection (55.0\% vs 35.0\% for DCHR and 85.5\% vs 82.0\% for ADODL) and Claude on the 2020 edition (47.1\% vs 31.4\% for DCHR and 83.9\% vs 80.9\% for ADODL).

\subsection{The Impact of the Adaptive $k^*$ Dense Retrieval Approach} \label{adaptkstars}

Figure \ref{kstars} shows how performance metrics change as we vary the number of retrieved documents $k$ in the eRisk 2019 dataset.
We analyze the performance of different metrics using \texttt{gpt-4o-mini} as the LLM and \texttt{MiniLM-L12-v3} as the retrieval model, while varying the parameter $k$ across \{5, 10, 15, 20, 30, 40, 50\}, where $k = 15$ represents the mean value of $k^*$. We test this scenario on both the 2019 and 2020 eRisk editions. We observe that using $k^*$ (horizontal dashed lines), which corresponds to adaptive RAG, often yields the best results.

For the 2019 edition, the best values are achieved with $k^*$ for the questionnaire metrics (DCHR, ADODL), while $k = 10$ produces the best metric in terms of AHR, and $k = 15$ performs best for ACR (item metrics) - both values being close to the mean $k^*$ value.

The 2020 edition shows slightly different results: for DCHR, the best value is obtained with the mean $k^*$, while the best ADODL and AHR corresponds to $k^*$. The best ACR results are obtained with $k = 30, 50$.

Overall, while the adaptive RAG strategy may not always lead to optimal results across all metrics, it allows us to achieve consistently strong performance without the need to set any parameters a priori or explore different choices of $k$. This automated approach to determining $k$ offers a robust and efficient solution that removes the need for manual parameter tuning while maintaining competitive performance levels across different evaluation scenarios and metrics.

\begin{table}[ht]
\small
\centering
\begin{tabular}{|l|c|c|c|c|}
\hline
\cellcolor{mygray}\textbf{LLMs} & \cellcolor{mygray}\textbf{DCHR} & \cellcolor{mygray}\textbf{ADODL} &\cellcolor{mygray} \textbf{AHR} & \cellcolor{mygray}\textbf{ACR}\\
\hline
\multicolumn{5}{|c|}{\cellcolor{mygray}$ \mu_{direct} > \mu_{CoT}$ }\\
\hline
\texttt{gpt-4o-mini}&  \cellcolor{lightindianred} & $\dagger$  & $\dagger$ & $\dagger$ \\
\hline
\texttt{Claude Sonnet} &\cellcolor{lightindianred}  &$\dagger$  & \cellcolor{lightindianred}& \cellcolor{lightindianred}\\
\hline
\texttt{Qwen 2.5 70B} & \cellcolor{lightindianred} & \cellcolor{lightindianred} & \cellcolor{lightindianred}&\cellcolor{lightindianred} \\
\hline
\texttt{DeepSeek V3} & \cellcolor{lightindianred} &  $\dagger$&\cellcolor{lightindianred} & \cellcolor{lightindianred}\\
\hline
\texttt{Phi-3-mini}& \cellcolor{lightindianred} & $\ddagger$ & $\ddagger$& $\ddagger$\\
\hline
\texttt{Phi-3.5-mini}& \cellcolor{lightindianred}& $\ddagger$ &$\ddagger$ & $\ddagger$ \\
\hline
\end{tabular}
\normalsize
\caption{The significance of the LLMs prediction goodness w.r.t. the two prompting techniques used, Direct and CoT, is shown (according to the eRisk 2019 collection). Metrics with no significant difference are marked in red, while \( \dagger \) denotes a significant difference according to the t-test, and \( \ddagger \) denotes significance according to the Mann-Whitney U test as well.}
\label{test}
\end{table}

\subsection{The Impact of Different Prompting Strategies}
In Table \ref{test}, we evaluate whether the use of direct prompting is statistically better than CoT across different metrics w.r.t. eRisk 2019 dataset collection. We perform both parametric, t-test, and non-parametric, Mann-Whitney U test (in both cases, we test whether one population mean is statistically greater than the other, $\alpha = 0.05$).
Lightweight open-source models (\texttt{Phi-3-mini} and \texttt{Phi-3.5-mini}) perform well only in direct prompt contexts, while they perform poorly when the prompting technique is CoT, tending to overestimate questionnaire scores (difference is statistically significant for ADODL, AHR, and ACR metrics).

For closed-source (\texttt{gpt-4o-mini}, \texttt{Claude-3.5-Sonnet}) and large-scale open-source LLMs (\texttt{Qwen 2.5 70B}, \texttt{DeepSeek V3}), the performance difference between prompting strategies varies. Specifically, \texttt{gpt-4o-mini} shows significant differences only under t-test across most metrics. \texttt{Claude-3.5-Sonnet} and \texttt{DeepSeek V3} demonstrate significant differences solely in ADODL according to t-test, while \texttt{Qwen 2.5 70B} shows no statistically significant differences across any metric.

\subsection{The Impact of Different Retrieval Approaches} 
\begin{figure}[t]
    \centering
    \includegraphics[height=.8\textwidth]{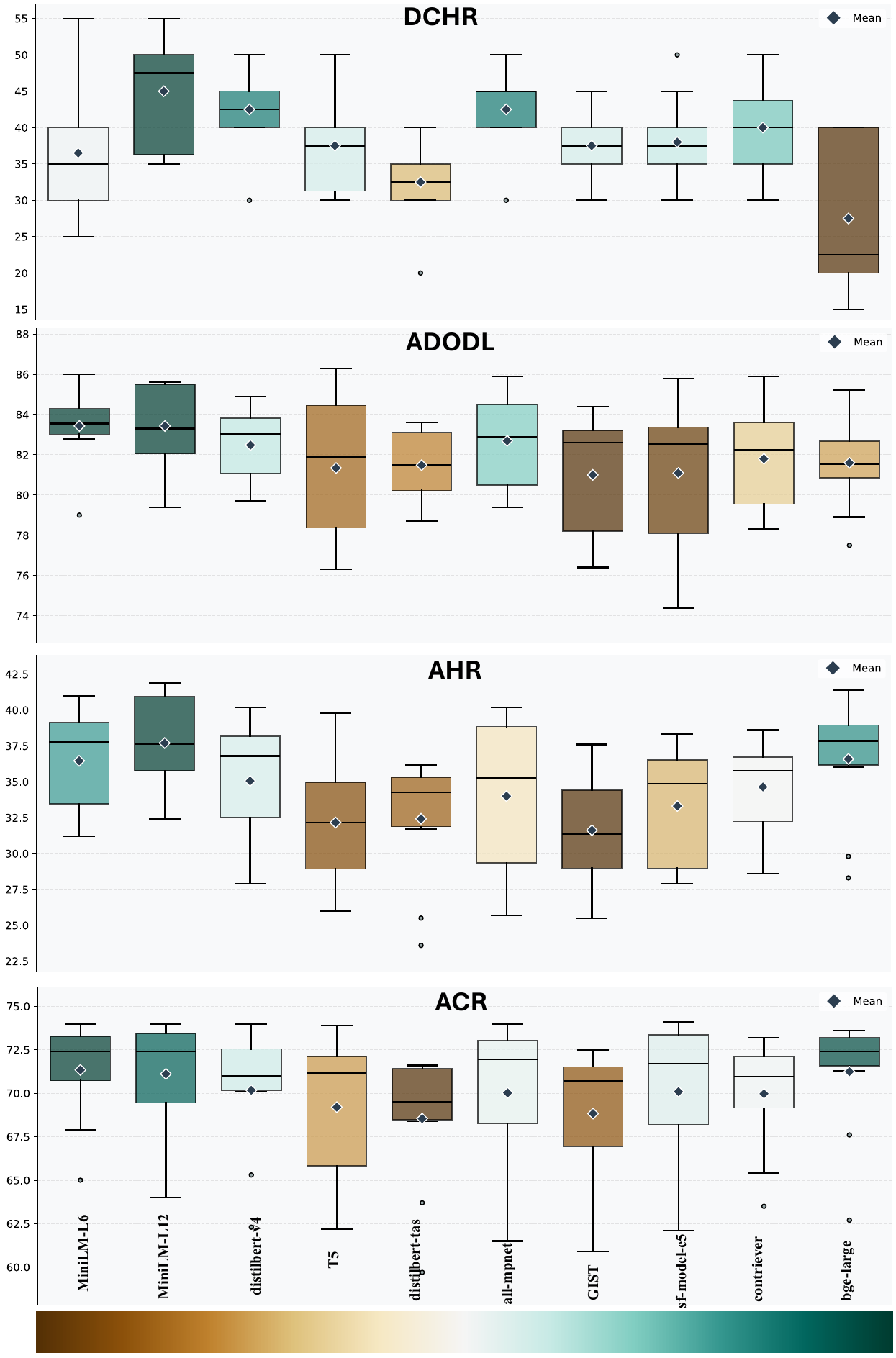}
    \caption{Distribution of adaptive RAG scores conditioned on different retrieval approaches and their rankings for each metric on the eRisk 2019 dataset. Rankings are based on mean scores (shown as diamonds). The color gradient from brown to teal indicates performance ranking, with darker teal representing better performance.}
    \label{boxs}
\end{figure}

We also observe that some dense retrieval models perform better globally compared to others (see Figure \ref{heat}), while some perform better only with respect to a subset of LLMs.
In Figure \ref{boxs}, we examine the distribution of LLMs scores across different retrieval approaches and their rankings for each metric w.r.t. eRisk 2019 dataset collection (after removing anomalous values given by lightweight open-source models in combination with CoT prompting strategies). While no single retrieval approach consistently outperforms the others across all metrics, we can observe that some retrievers (\textit{msmarco-MiniLM-L-6-v3}, \textit{msmarco-MiniLM-L-12-v3}, \textit{distillbert-v4}, and \textit{bge-large}) generally achieve better rankings on average.

\end{document}